\documentclass[11pt]{article}
\usepackage[utf8]{inputenc}
\usepackage{coling2016}
\usepackage{amsmath}
\usepackage{amssymb}
\usepackage{times}
\usepackage{url}
\usepackage{latexsym}
\usepackage{graphicx}
\usepackage{booktabs}
\usepackage{multirow}
\usepackage{array}
\usepackage{rotating}

\usepackage[usenames, dvipsnames]{color}

\newcommand\ie[1]{\textit{i.e.~}}
\newcommand\eg[1]{\textit{e.g.~}}

\title{Multimodal Attention for Neural Machine Translation}

\author{Ozan Caglayan$^{1,2}$,\quad Lo\"ic Barrault$^1$,\quad Fethi Bougares$^1$ \\
$^1$ LIUM, University of Le Mans / France\\
$^2$ Galatasaray University / Turkey\\
    {\tt $^1$FirstName.LastName@univ-lemans.fr} \\
    {\tt $^2$ocaglayan@gsu.edu.tr}}

\date{}

\begin{document}
\maketitle
\begin{abstract}
The attention mechanism is an important part of the neural machine translation (NMT) where it was reported to produce
richer source representation compared to fixed-length encoding sequence-to-sequence models.
Recently, the effectiveness of attention has also been explored in the context of image captioning.
In this work, we assess the feasibility of a multimodal attention mechanism that simultaneously focus over an image and its natural
language description for generating a description in another language.
We train several variants of our proposed attention mechanism on the Multi30k multilingual image captioning dataset. We show that a
dedicated attention for each modality achieves up to 1.6 points in BLEU and METEOR compared to a textual NMT baseline.
\end{abstract}


\section{Introduction}
\label{intro}
While dealing with multimodal stimuli in order to perceive the surrounding environment and to understand the world is natural for human beings,
it is not the case for artificial intelligence systems where an efficient integration of multimodal and/or multilingual information
still remains a challenging task. Once tractable, this grounding of multiple modalities against each other may enable a more natural language
interaction with computers.

Recently, deep neural networks (DNN) achieved state-of-the-art results in numerous tasks of computer vision (CV), natural language processing (NLP) and speech processing
\cite{lecun2015deep} where the input signals are monomodal i.e. image/video, characters/words/phrases, audio signal. With these successes in mind,
researchers now attempt to design systems that will benefit from the fusion of several modalities in order to reach a wider generalization
ability.

Machine translation (MT) is another field where purely neural approaches (NMT) \cite{Sutskever2014,Bahdanau2014}
challenge the classical phrase-based approach \cite{koehn2003statistical}. This has been possible by formulating the
translation problem as a sequence-to-sequence paradigm where a DNN may read and produce text in discrete steps with
special recurrent building blocks that can model long range dependencies.

Another straightforward application of this paradigm is to incorporate DNNs into the task of generating natural language descriptions
from images, a task commonly referred as image captioning. A number of approaches proposed by \newcite{karpathy2015deep} ,\newcite{vinyals2015show}, \newcite{xu2015show}, \newcite{you2016image}
have achieved state-of-the-art results on popular image captioning datasets Flickr30k \cite{young2014image} and MSCOCO \cite{lin2014microsoft}.

In this paper, we propose a multimodal architecture that is able to attend to multiple input modalities and we
realize an extrinsic analysis of this specific attention mechanism. Furthermore, we attempt to determine
the optimal way of attending to multiple input modalities in the context of image captioning.
Although several attempts have been made to incorporate features from different languages (multi-source/multi-target) or different tasks
to improve the performance \cite{Elliott2015,Dong2015MultiTaskLF,luong2015multi,firat2016multi}, according to our knowledge, there has not been any attentional neural translation or captioning approach trained using an auxiliary source modality.

The architecture that we introduce for achieving multimodal learning can either be seen as
an NMT enriched with convolutional image features as auxiliary source representation
or an image captioning system producing image descriptions in a language T, supported with source descriptions in another language S.
The modality specific pathways of the proposed architecture are inspired from two previously published approaches \cite{Bahdanau2014,xu2015show} where
an attention mechanism is learned to focus on different parts of the input sentence.
Such integration implies a careful design of the underlying blocks: many strategies are considered in order to correctly attend to multiple input modalities. Along this line, different possibilities are also assessed for taking into account multimodal information during the decoding process of target words.

We compare the proposed architecture against single modality baseline systems using the recently published Multi30k multilingual image captioning dataset \cite{Elliott2016}.
The empirical results show that we obtain better captioning performance in terms of different evaluation metrics when compared to baselines.
We further provide an analysis of the descriptions and the multimodal attention mechanism to examine the impact of the multimodality
on the quality of generated descriptions and the attention.

\section{Related Work}
\label{related}
Since the presented work is at the intersection of several research topics, this part is divided into further sections for clarity.

\subsection{Neural Machine Translation (NMT)}
End-to-end machine translation using deep learning has been first proposed by \newcite{kalchbrenner2013}
and further extended by \newcite{cho2014gru}, \newcite{Sutskever2014} and \newcite{Bahdanau2014} to achieve competitive results compared to state-of-the-art phrase-based method \cite{koehn2003statistical}. The dominant approaches in NMT differ in the way of representing the source sentence: \newcite{Sutskever2014} and \newcite{cho2014gru} used the last hidden state of the encoder as a fixed size source sentence representation while \newcite{Bahdanau2014} introduced an attention mechanism where a set of attentional weights are assigned to the recurrent hidden states of the encoder to obtain a weighted sum of the states instead of just taking the last. The idea of attention still continues to be an active area of research in the NMT community \cite{luong2015effective}.

\subsection{Neural Image Captioning}
End-to-end neural image description models are basically recurrent language models conditioned in different ways on image features. These image features are generally extracted from powerful state-of-the-art CNN architectures trained on large scale image classification tasks like ImageNet \cite{deng2009imagenet}. \newcite{mao2014deep}, \newcite{karpathy2015deep}, \newcite{vinyals2015show} proposed a multimodal RNN which differs in the selection and integration of image features: \newcite{mao2014deep} made use of a multimodal layer which fuses image features, current word embedding and current hidden state of the RNN into a common multimodal space. The image features experimented by the authors are extracted from two different CNNs namely AlexNet \cite{krizhevsky2012imagenet} and VGG \cite{simonyan2014very}. \newcite{karpathy2015deep} takes a simpler approach and used a vanilla RNN that incorporates VGG image features only at the first time step as a bias term. Finally \newcite{vinyals2015show} trained an ensemble of LSTM in which the image features extracted from a batch-normalized GoogLeNet \cite{ioffe2015batch} are presented to the LSTM sentence generator as the first input, before the special start word.

Different from the previously cited works, \newcite{xu2015show} applied the attention mechanism over convolutional image features of size 14x14x512 extracted from VGG, which makes the image context a collection of feature maps instead of a single vector. During training, an LSTM network jointly learns to generate an image description while selectively attending to the presented image features. This model is similar to the attentional NMT introduced by \newcite{Bahdanau2014} except that the source word annotations produced by the encoder are replaced by the convolutional features.

Another attention based model proposed by \newcite{you2016image} introduced two separate attention mechanisms, input and output attention models, that are applied over a set of visual attributes detected using different methods like k-NN and neural networks. The image representation extracted from GoogLeNet \cite{szegedy2015going} is only used as the initial input to the RNN. The authors use an ensemble of 5 differently initialized models as their final system.

In the context of multimodality, \newcite{Elliott2015} explore the effectiveness of conditioning a target language model on the image features from the last fully-connected layer of VGG and on the features from a source language model using the IAPR-TC12 multilingual image captioning dataset. 

\subsection{Multi-Task Learning}
Several recent studies in the literature explored multi-task learning for different NLP and MT
tasks. \newcite{Dong2015MultiTaskLF} proposed a multi-target NMT that translates a single source language
into several target languages. Specifically, they made use of a single encoder and multiple
language-specific decoders each embedded with its own attention mechanism.
\newcite{firat2016multi} extended this idea into a multi-way multilingual NMT which can translate
between multiple languages using a single attention mechanism shared across all the languages.
\newcite{luong2015multi} experimented with \textit{one-to-many, many-to-one and many-to-many}
schemes in order to quantify the mutual benefit of several NLP, MT and CV tasks to each other.

\section{Architecture}
\label{architecture}
\begin{figure*}[t]
\centering
  \includegraphics[width=.94\textwidth]{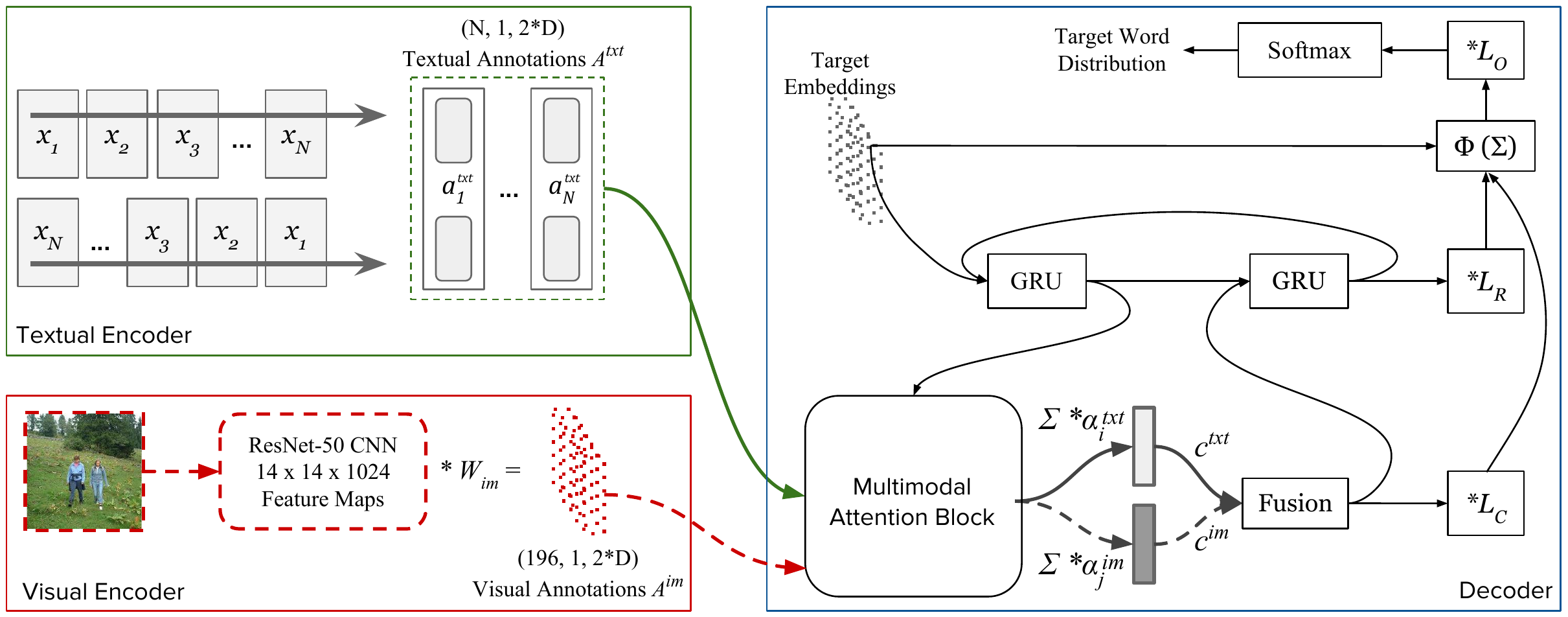}
  \caption{The architecture of MNMT: The boxes with $*$ refer to a linear transformation while $\Phi(\Sigma)$ means a $tanh$ applied over the sum of the inputs.}
  \label{fig:mnmt_arch}
\end{figure*}

\newcommand{\cev}[1]{\reflectbox{\ensuremath{\vec{\reflectbox{\ensuremath{#1}}}}}}
The general architecture (Figure~\ref{fig:mnmt_arch}) is composed of four building blocks namely the textual encoder, the visual encoder, the attention mechanism and the decoder, which will all be described in what follows.

\subsection{Textual Encoder}
We define by $X$ and $Y$, a source sentence of length $N$ and a target sentence of length $M$ respectively where each word is represented in a source/target embedding space with dimension $E$:
\begin{gather}
    X = (x_{1},x_{2}, ... ,x_{N}), x_{i} \in{\mathbb{R}^{E}} \\
    Y = (y_{1},y_{2}, ... ,y_{M}), y_{j} \in{\mathbb{R}^{E}}
\end{gather}
A bi-directional GRU \cite{cho2014gru} encoder with $D$ hidden units reads the input $X$ sequentially in forwards and backwards to produce two sets of hidden states based on the \textit{current} source word embedding and the \textit{previous} hidden state of the encoder.
We call $a^{txt}_{i}$ the textual annotation vector obtained at time step $i$ by concatenating the forward and backward hidden states of the encoder:
\begin{equation}
a^{txt}_{i} = \begin{bmatrix}
	\vec{h_{i}} \\
  \cev{h_{i}}
  \end{bmatrix}, a^{txt}_{i} \in{\mathbb{R}^{2D}}
\end{equation}
At the end of this encoding stage, we obtain a set of $N$ textual annotation vectors $A^{txt}=\{a^{txt}_{1},a^{txt}_{2},\dotsc,a^{txt}_{N}\}$ that represents the source sentence.

\subsection{Visual Encoder}
We use convolutional feature maps of size 14x14x1024 extracted from the so-called \emph{res4f\_relu} layer (end of Block-4, after ReLU) of a ResNet-50 trained on ImageNet (See Section~\ref{dataset} for details).
In order to make the dimensionality compatible with the textual annotation vectors $a^{txt}_{i}$, a linear transformation $W_{im}$ is applied to the image features leading to the visual annotation vectors $A^{im}=\{a^{im}_1,a^{im}_{2},\dotsc,a^{im}_{196}\}$.

\subsection{Decoder}
A conditional GRU\footnote{\texttt{github.com/nyu-dl/dl4mt-tutorial/blob/master/docs/cgru.pdf}} (CGRU) decoder has been extended for multimodal context and equipped with a multimodal attention mechanism in order to generate the description in the target language.
CGRU consists of two stacked GRU activations that we will name $g_1$ and $g_2$ respectively.
We experimented with several ways of initializing the hidden state of the decoder and found out that the model performs better when $h^{(1)}_{0}$ is learned from the mean textual annotation using a feed-forward layer with $tanh$ non-linearity. The hidden state $h^{(2)}$ of $g_2$ is initialized with $h^{(1)}_{1}$:
\begin{equation}
  h^{(1)}_{0} = tanh\left(W^{T}_{init} \left( \frac{1}{N}\sum_{i=1}^{N} a^{txt}_{i} \right) + b_{init} \right) \quad , \quad
  h^{(2)}_{0} = h^{(1)}_{1}
\end{equation}
At each time step, a multimodal attention mechanism (detailed in Section~\ref{attention}) computes two modality specific context vectors $\{c^{txt}_t, c^{im}_t\}$ given the hidden state of the first GRU $h^{(1)}_{t}$ and the textual/visual annotations $\{A^{txt}, A^{im}\}$.
The multimodal context vector $c_t$ is then obtained by applying a $tanh$ non-linearity over a fusion of these modality specific context vectors.
We experimented with two different fusion techniques that we will refer to as SUM($F_S$) and CONCAT($F_C$):
\begin{align}
  c_t = F_{S}(c^{txt}_t, c^{im}_t) &= \tanh(c^{txt}_t + c^{im}_t)\\
  c_t = F_{C}(c^{txt}_t, c^{im}_t) &= \tanh \left( W^{T}_{fus}\begin{bmatrix} c^{txt}_t \\[.4em] c^{im}_t \end{bmatrix} + b_{fus} \right) \label{eq:fusion}
\end{align}

The probability distribution over the target vocabulary is conditioned on the hidden state $h^{(2)}_{t}$ of the second GRU (which receives as input $c_t$),
the multimodal context vector $c_t$ and the embedding of the (true) previous target word $E_{y_{t-1}}$:
\begin{align}
  h^{(2)}_{t} &= g_2(h^{(2)}_{t-1}, c_t)\\
  P(y_t=k|y_{<t}, A^{txt}, A^{im}) &= softmax\left(L_o \tanh(L_s{h^{(2)}_t} + L_c{c_t} + E_{y_{t-1}})\right) \label{eq:generate}
\end{align}

\section{Attention Mechanism}
\label{attention}
In the context of multi-way multilingual NMT, \newcite{firat2016multi} benefit
from a single attention mechanism shared across different language pairs.
Although this may be reasonable within a multilingual task where
all inputs and outputs are solely based on textual representations,
this may not be the optimal approach when the modalities are completely
different.

\begin{figure}[t!]
\begin{center}
\includegraphics[width=.9\textwidth]{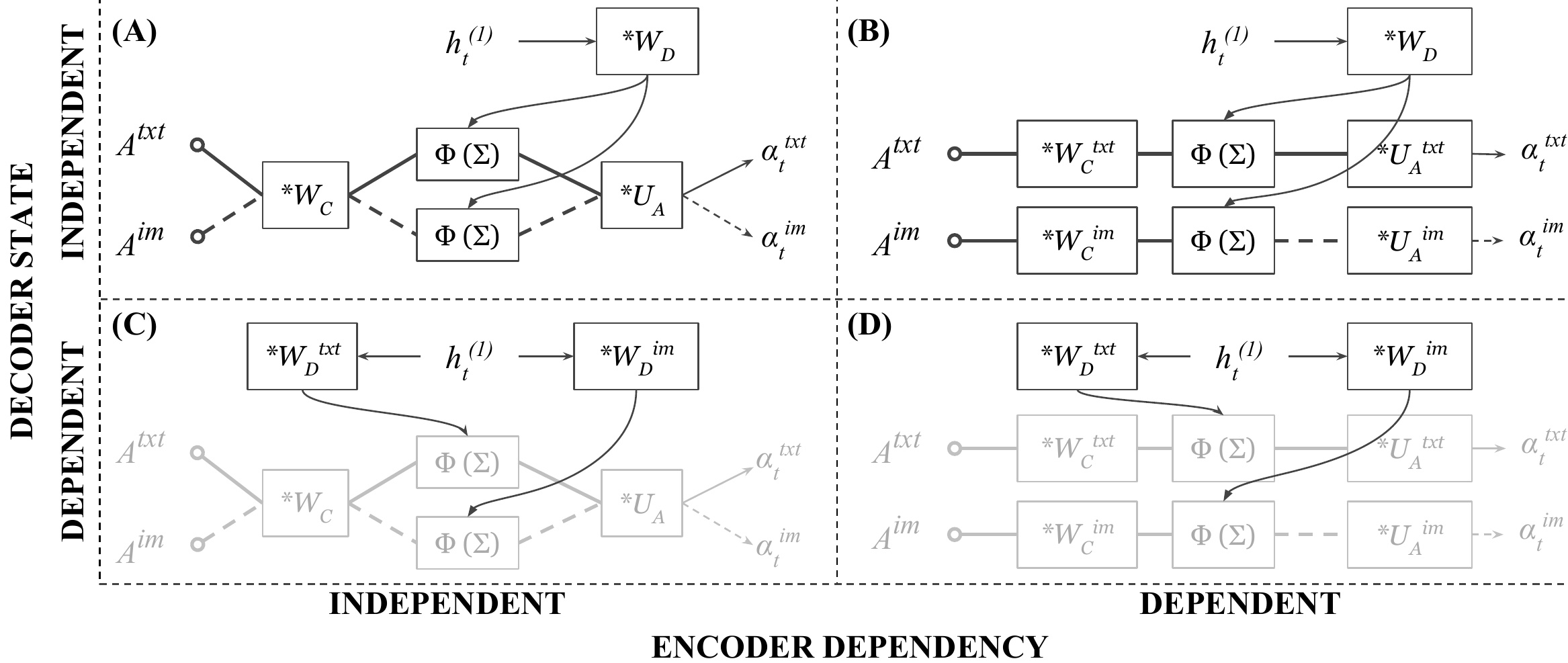}
  \caption{The conceptualization of multimodal attention in terms of different dependency schemes over the source modalities. Common parts from (A) to (C) and (B) to (D) are grayed out to emphasize the changes.}
\label{fig:attention}
\end{center}
\end{figure}

In order to verify the above statement, we consider different schemes for integrating the attention mechanism into multimodal learning.
First, a shared feed-forward network is used to produce modality-specific attention weights $\{\alpha^{txt}_t,\alpha^{im}_t\}$:
\begin{align}
  \alpha^{txt}_t &= softmax\left(U_A\tanh(W_D\ h^{(1)}_t + W_C\ A^{txt})\right)\\
  \alpha^{im}_t  &= softmax\left(U_A\tanh(W_D\ h^{(1)}_t + W_C\ A^{im}) \right)
\end{align}
We call this basic approach where linear transformations $\{W_C,U_A\}$ from contexts to attentional weights are all shared, the \textit{encoder-independent attention} (Figure~\ref{fig:attention}A).
An \textit{encoder-dependent attention} in contrast, has distinct $\{W_C,U_A\}$ for each modality which means that they are projected differently (Figure~\ref{fig:attention}B).

It should be noted that both mechanisms depicted above still have a common projection layer $W_D$ applied
to the hidden state $h^{(1)}_t$ of the first GRU. In order to objectively assess the impact of
modality-dependency on the decoder side, we also need to consider the two cases where a shared $W_D$ or distinct $\{W^{txt}_D,W^{im}_D\}$
transformations are used in the projection layer (Figure ~\ref{fig:attention}C and ~\ref{fig:attention}D).
These two variants will be respectively referred as \textit{independent} and \textit{dependent decoder state projection} in the rest of the paper.

To sum up, we propose four variants of the multimodal attention mechanism in terms of modality dependency with respect to encoder and decoder.
While encoder-independent attention with dependent decoder state projection
(Figure~\ref{fig:attention}C) might seem unnatural for this single target task, it may be
interesting as a contrastive system.

Once the attentional weights $\{\alpha^{txt}_t,\alpha^{im}_t\}$ are computed, the modality-specific context vectors are defined as the weighted sum of each annotation set with its relevant $\alpha$:
\begin{align}
  c^{txt}_t = \sum_{i=1}^{N}  \alpha^{txt}_{ti}\ a^{txt}_{i}\quad,\quad
  c^{im}_t  = \sum_{j=1}^{196} \alpha^{im}_{tj}\ a^{im}_{j}
\end{align}
These vectors are then merged into a multimodal context vector that will be used in the target word generation (see equations~\ref{eq:fusion}, \ref{eq:generate} in Section~\ref{architecture}).

\section{Experiments}
\label{experiments}
\subsection{Dataset}
\label{dataset}

We used the Multi30K dataset \cite{Elliott2016} which is an extended version of the Flickr30K Entities dataset \cite{young2014image}. Multi30K extends the original Flickr30K that contains 31K images with 5 English descriptions with 5 more independently crowdsourced descriptions in German. It should be emphasized that the provided bilingual descriptions are \textit{not} direct translations but could be considered somewhat comparable.

The training, validation and test set consist of 29000, 1014 and 1000 images respectively each having 5 English and 5 German annotations. A total of 25 sentence pairs can be obtained by taking the cross product of these annotations leading to a training set of 725K samples. Throughout this work, we reduced 725K to 145K by considering only 5 sentence pairs for each image.


\begin{table}[ht]
\centering
\resizebox{0.4\columnwidth}{!}{%
\begin{tabular}{@{}lccr@{}}
\toprule
Side    & Vocabulary & \# Words & Avg. Length\\ \midrule
English & 16802      & 1.5M & 11.7 words \\
German  & 10000      & 1.3M & 9.8 words
\end{tabular}}
\caption{Summary of the training data.}
\label{tbl:train_data}
\end{table}

We tokenized the sentences with \texttt{tokenizer.perl} from Moses, removed punctuation and lowercased the sentences. We only kept sentence pairs with sentence lengths $\in{[3,50]}$ having a length ratio of at most 3. This results in a final training dataset of 131K sentences (Table \ref{tbl:train_data}). We picked the most frequent 10K German words and replaced the rest with an UNK token for the target side.

For the image part, convolutional image features of size 14x14x2014 are extracted from the \textbf{res4f\_relu} layer of ResNet-50 CNN \cite{he2016resnet} trained on ImageNet. The images were resized to 224x224 without any cropping prior to extraction. The final features are used as 196x1024 dimension matrices per each image during training and testing.

\subsection{Training}
We trained two monomodal baselines namely NMT, IMGTXT and different attentional variants of our proposed MNMT. The baselines have exactly the same architecture presented throughout this work but with only a single source modality.
All models have word embeddings and recurrent layers of dimensionality 620 and 1000 respectively. We used Adam \cite{kingma2014adam} as the stochastic gradient descent variant with a minibatch size of 32. The weights of the networks are initialized using Xavier scheme \cite{glorotxavier} while the biases are initially set to 0. L2 regularization with $\lambda=0.00001$ is applied to the training cost to avoid overfitting.

The performance of the networks is evaluated on the first validation split using BLEU \cite{bleu} at the end of each epoch and the training is stopped if BLEU does not improve for 20 evaluation periods.

A classical left to right beam-search with a beam size of 12 is used for sentence generation during test time. Besides evaluating performance
on the first validation split, we also experimented with a best source selection strategy where for each image we obtain 5 German hypotheses that
correspond to 5 English descriptions and pick the one with the highest log-likelihood and the least number of UNK tokens.

\section{Results}
\label{results}
\newcommand{\sss}[1] {\textsuperscript{#1}}
\newcommand{\ig}[1] {\includegraphics[scale=0.4]{figures/#1.jpg}}
\newcommand{\vt}[1] {\vspace{6em}\begin{sideways}\textbf{#1}\end{sideways}}
\newcommand{\tb}[1] {\textbf{#1}}

\newcolumntype{L}{>{\raggedright\let\newline\\\arraybackslash\hspace{0pt}}m{4.6cm}}
\newcolumntype{T}{>{\centering\let\newline\\\arraybackslash\hspace{0pt}}m{0.1cm}}
\newcolumntype{C}{>{\centering\let\newline\\\arraybackslash\hspace{0pt}}m{3.7cm}}
\newcolumntype{R}{>{\raggedright\let\newline\\\arraybackslash\hspace{0pt}}m{4.2cm}}

\begin{table}[h]
\centering
\resizebox{0.7\columnwidth}{!}{%
\begin{tabular}{@{}rrcclll@{}}
\toprule
\multicolumn{1}{c}{} & {} & \multicolumn{2}{c}{Attention Type} & \multicolumn{3}{c}{Validation Scores}           \\ \midrule
\multicolumn{1}{c}{Model} & Fusion & Modality & Decoder & METEOR   & BLEU               & CIDEr-D            \\ \midrule
  NMT    &  	-	  & 	-	      &  	-	   	  & 34.24 (35.59)         & 18.64 (21.62)      & 58.57 (67.93)      \\
  IMGTXT &  	-	  & 	-	      &  	-	   	  & 26.80                 & 11.16              & 31.28              \\ \midrule
  MNMT1   & SUM 	  & IND       & IND   	  & 33.23 (35.42)         & 18.30 (21.24)      & 55.45 (65.03)      \\
  MNMT2  & SUM 	  & IND       & DEP       & 34.17 (35.48)         & 17.70 (20.70)      & 53.78 (61.76)      \\
  MNMT3   & SUM 	  & DEP       & IND   	  & 34.38 (35.55)         & 18.42 (20.94)      & 55.81 (63.37)      \\
  MNMT4   & SUM 	  & DEP       & DEP       & 33.67 (34.57)         & 17.83 (20.30)      & 52.68 (59.63)      \\ \midrule
  MNMT5   & CONCAT	& IND       & IND   	  & 33.31 (34.98)         & 17.50 (20.60)      & 53.57 (61.46)      \\
  MNMT6   & CONCAT	& IND       & DEP       & \tb{35.23} (36.79)    & 19.30 (22.45)      & 60.62 (69.96)      \\
  MNMT7   & CONCAT	& DEP       & IND       & \tb{35.11 (37.13)}    & \tb{19.72 (23.24)} & \tb{61.04 (72.16)} \\
  MNMT8   & CONCAT	& DEP       & DEP       & 34.80 (\tb{36.98})         & 19.55 (22.78)      & 60.20 (70.20)      \\ \bottomrule
\end{tabular}}
  \caption{The results for the first validation split and the best source selection (between parentheses). All scores are averages of two runs.}
\label{tbl:results1}
\end{table}

\subsection{Quantitative Analysis}
The description generation performance of the models is presented in Table~\ref{tbl:results1} using BLEU, METEOR \cite{meteor} and CIDEr-D \cite{vedantam2015cider} as automatic evaluation metrics. 

It is clear from Table~\ref{tbl:results1} that MNMT with CONCAT as the fusion operator improves over both the NMT and the IMGTXT baselines when combined with modality-dependent attention mechanism (MNMT 6 to 8). 
The results are slightly worse than the NMT baseline regardless of the multimodal attention type when the fusion is realized with the SUM operator. This difference can be attributed to the fact that concatenation makes use of a linear layer that learns how to integrate the modality-specific activations into the multimodal context vector.

The improvement in terms of the automatic metrics is more significant with the best source selection strategy compared to the first validation split. The reason for this is that the first split contains source sentences which are much longer and detailed than the target ones: 18.35 words/sentence in average on the source side compared to 8.03 words/sentence on the target side. Once we use the best source selection method, the models compensate for this discrepancy by choosing source sentences of different lengths (in average $\sim$13 words/sentence).

We observe that a completely independent (shared) attention mechanism (MNMT5) has the worst performance among all CONCAT variants.
This empirical evidence is on par with our initial statement that a single shared attention may not be the optimal approach in the case of different input modalities.
We also notice that once a dependency be it on the encoder or the decoder side is established in the topology,
the performance improves compared to the NMT baseline and MNMT5.
The results obtained by MNMT6 that we previously conjectured as unnatural (see Section~\ref{attention}) also confirm this.
\subsection{Qualitative Analysis}
\begin{table}[h]
  \scriptsize
\centering
\begin{tabular}[t]{TrLLR}
  \multicolumn{2}{T}{}              & \multicolumn{1}{C}{\ig{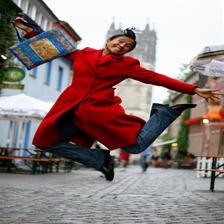}} & \multicolumn{1}{C}{\ig{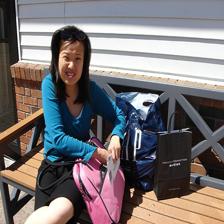}} & \multicolumn{1}{C}{\ig{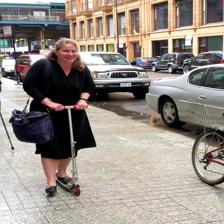}} \\ \midrule
  \multicolumn{2}{T}{\vspace{1em}Source}    & \multicolumn{1}{L}{a woman in jeans and a red coat and carrying a multicolored handbag spreads put her arms while leaping up from a cobblestone street}
                                  & \multicolumn{1}{L}{an asian woman sitting on a bench going through a pink laptop bag}
                                  & \multicolumn{1}{R}{women in a black dress riding a scooter down the street} \\ \midrule

  \multirow{5}{*}{\vt{NMT}} & HYP & \multicolumn{1}{L}{eine frau springt auf einem gehweg in die luft (47.84)} & {eine frau sitzt auf einer bank und hält einen laptop in der hand (52.61)} & {eine frau in weißem kleid fährt auf einem roller eine straße entlang (44.00)}                    \\ \cmidrule(l){2-5}
                     & ENG & \multicolumn{1}{L}{\textit{a woman jumps on a walkway in the air}} & {\textit{a woman sitting on a bench and \tb{holding a laptop} in hand}} & {\textit{a woman in a \tb{white} dress riding a scooter a road along}} \\ \midrule
  \multirow{5}{*}{\vt{MNMT}} & HYP & \multicolumn{1}{L}{eine frau in rotem anorak und schwarzer hose springt mit ausgebreiteten armen durch die luft (49.09)} & {eine asiatische frau sitzt auf einer holzbank (34.01)} & {eine frau im schwarzen kleid auf einem motorroller (31.15)}                           \\ \cmidrule(l){2-5}
                     & ENG & \multicolumn{1}{L}{\textit{a woman in a \tb{red suit and black trousers} jumping with \tb{outstretched arms} through the air}}     & {\textit{an \tb{asian} woman sitting on a \tb{wooden bench}}} & {\textit{a woman in \tb{black} dress on a scooter}} \\ \midrule
\end{tabular}
\caption{Comparison of NMT and MNMT models on the first validation split: ENG is the Google Translate output of the German hypotheses and the values in parentheses are METEOR sentence scores.}
\label{tbl:comparison}
\end{table}
In Table~\ref{tbl:comparison} we present German descriptions generated by the NMT baseline and the best performing MNMT model using the same source description from the first validation split. In the first image, MNMT clearly produces a richer description where it provides additional visual information like the color and the type of clothing and also the positioning of the woman. In the second case, MNMT again generates a coherent and rich description but ignores the \textit{pink laptop bag} that is mentioned in the NMT output. Lastly, third image shows a situation where NMT wrongly describes the color of an object while MNMT does its job correctly although METEOR penalizes it, a phenomenon observed when the references contain less detail than the hypotheses.

\begin{figure*}[hbtp!]
\centering
\includegraphics[width=.99\textwidth]{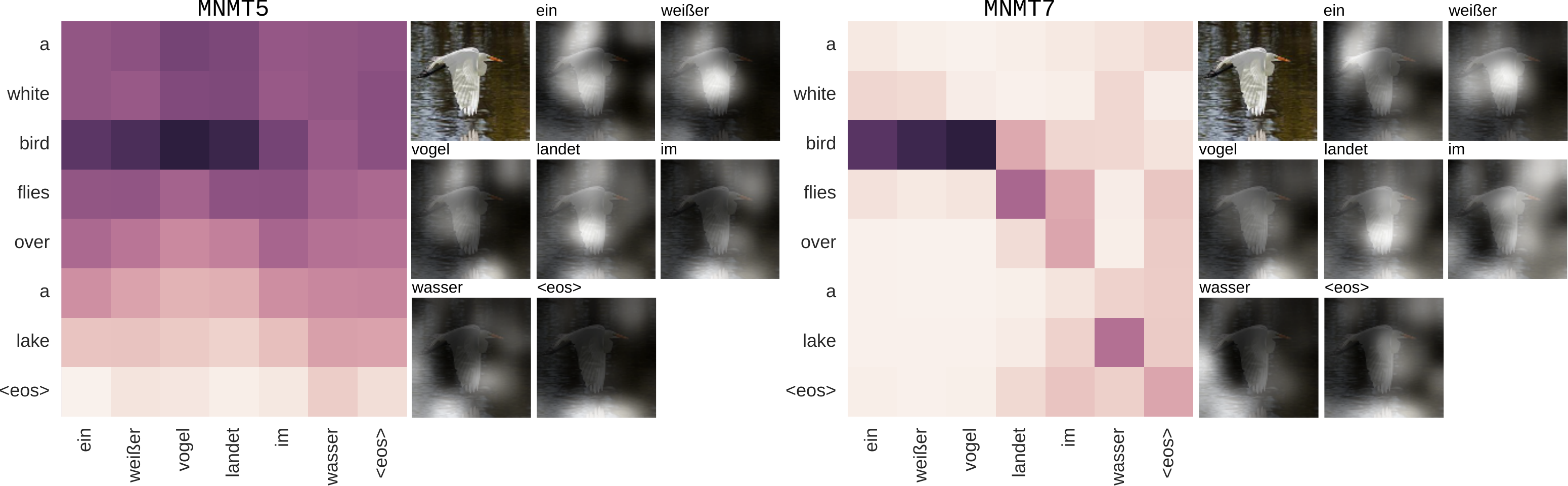}
  \caption{The impact of sharing the attention on the attention precision: (Left) Completely independent (shared) attention. (Right) Encoder-dependent attention with independent decoder state projection. (The description in English: \textit{a white bird landing in water})}
\label{fig:att_comparison}
\end{figure*}

One advantage of the attention mechanism is the ability to visualize where exactly the network pays attention. Although the visualization over the textual context is trivial to apply, inferring the spatial region in the original image that is attended by the model is not. Here we adopt the approach proposed by \newcite{xu2015show} which upsamples the attention weights with a factor of 16 in order to lay them over the original image as white smoothed regions.
We compare the multimodal attention between completely independent MNMT5 and encoder-dependent MNMT7 in Figure~\ref{fig:att_comparison}.

Although both models achieve a comparable attention pattern over the image regions, textual alignment is completely
disturbed (for all examples) in MNMT5. The failure of the shared attention
which contrasts with \newcite{firat2016multi} can be attributed
to a representational discrepancy between convolutional image features and the source side textual annotations.
Moreover, the number of visual annotations for a single image (196) is $\sim$15 times higher than the number of
textual annotations (with 13 words/sentence in average) which may bias the joint attention towards the visual side during the learning stage.

\section{Conclusion}
\label{conclusion}
In this paper, we have proposed an architecture that performs multimodal machine translation with multimodal attention.
By quantitatively and qualitatively analyzing different attention schemes, we demonstrated that \textit{modality-dependent} MNMT outperforms the textual NMT baseline by up to 1 BLEU/METEOR and 2.5 CIDEr-D points.
The difference is even more significant with the best source selection strategy, where we reach a gain of almost
1.6 BLEU/METEOR and 4.2 CIDEr-D points. We further visualize multimodal attention and show that \textit{modality-dependent} attention mechanism is able to learn an alignment over both the source words and the image features but this is not the case for shared attention where textual
alignment is completely disturbed.

As future work, fine-tuning the extracted image features with a learnable convolutional layer
inside MNMT and/or extracting image features based on source language description are interesting
paths to explore in order to gain more insight about multimodality.

\bibliographystyle{acl}
\bibliography{nmt}

\end{document}